\documentclass{article} 
\usepackage{iclr2021_conference,times}

\usepackage{amsmath,amsfonts,bm}

\def\eqref#1{equation~\ref{#1}}

\def\1{\bm{1}}

\DeclareMathAlphabet{\mathsfit}{\encodingdefault}{\sfdefault}{m}{sl}
\SetMathAlphabet{\mathsfit}{bold}{\encodingdefault}{\sfdefault}{bx}{n}

\DeclareMathOperator*{\argmin}{arg\,min}

\usepackage{hyperref}
\usepackage{url}

\usepackage{booktabs}       
\usepackage{amsfonts}       
\usepackage{nicefrac}       
\usepackage{microtype}      

\usepackage{graphicx, wrapfig}
\usepackage{subcaption, floatrow}

\usepackage{xcolor}
\usepackage{amsmath}
\usepackage{hyperref}
\usepackage{amssymb}
\usepackage{cleveref}

\usepackage{algorithmic}

\usepackage{algorithm}

\usepackage{changepage}

\usepackage{makecell}

\renewcommand{\algorithmiccomment}[1]{{\color{blue}\texttt{\bgroup\hfill//~#1\egroup}}}

\newlength{\minuslength}
\newcommand*{\tran}{^{\mkern-1.5mu\mathsf{T}}}

\title{Regularization Can Help Mitigate Poisoning Attacks\ldots with the Right Hyperparameters}

\author{Javier Carnerero-Cano$^1$, Luis Mu\~noz-Gonz\'alez$^1$, Phillippa Spencer$^2$, Emil C. Lupu$^1$  \\
$^1$Department of Computing, Imperial College London\\
180 Queen's Gate, SW7 2AZ, London, United Kingdom \\
$^2$Cyber and Information Systems Division, Defence Science and Technology laboratory (Dstl)\\
Porton Down, Salisbury, United Kingdom\\

\texttt{\{j.cano, l.munoz, e.c.lupu\}@imperial.ac.uk}
}

\iclrfinalcopy 
\begin{document}

\maketitle

\begin{abstract}
 Machine learning algorithms are vulnerable to poisoning attacks, where a fraction of the training data is manipulated to degrade the algorithms' performance. We show that current approaches, which typically assume that regularization hyperparameters remain constant, lead to an overly pessimistic view of the algorithms' robustness and of the impact of regularization. We propose a novel optimal attack formulation that considers the effect of the attack on the hyperparameters, modelling the attack as a \emph{minimax bilevel optimization problem}. This allows to formulate optimal attacks, select hyperparameters and evaluate robustness under worst case conditions. We apply this formulation to logistic regression using $L_2$ regularization, empirically show the limitations of previous strategies and evidence the benefits of using $L_2$ regularization to dampen the effect of poisoning attacks.
\end{abstract}

\section{Introduction}
\label{sec:intro}
In many applications, Machine Learning (ML) systems rely for their training on data collected from \textit{untrusted} sources, such as humans, sensors, or IoT devices that can be easily compromised. Spoofed data from such sources can be used to poison the learning algorithms and maximize their error in targeted or indiscriminate ways. 
The vulnerability of ML algorithms to poisoning attacks has been amply studied \citep{biggio2012poisoning, mei2015using, xiao2015feature, koh2017understanding, munoz2017towards,  steinhardt2017certified, koh2018stronger, paudice2018detection,  shafahi2018poison, zhang2018training,  diakonikolas2019sever, munoz2019poisoning, zhu2019transferable, huang2020metapoison, geiping2021witches, levine2021deep}, as well as how adversaries can manipulate a fraction of the training data to subvert learning, decrease its overall performance or produce specific errors \citep{barreno2010security, huang2011adversarial, munoz2019challenges}. 

Several systematic poisoning attacks have been proposed to analyze, under worst-case conditions, different families of ML algorithms such as Support Vector Machines (SVMs) \citep{biggio2012poisoning}, other linear classifiers \citep{xiao2015feature, mei2015using, koh2018stronger}, and neural networks \citep{koh2017understanding, munoz2017towards, huang2020metapoison}. These attacks are typically formulated as a bilevel optimization problem where the attacker aims to maximize some malicious objective function (e.g. to maximize the error) by manipulating a fraction of the training data. At the same time, the defender aims to optimize a different objective function to learn the model's parameters, typically evaluated on the poisoned training set.

Some of these attacks target algorithms that have hyperparameters, but consider them constant regardless of the fraction of poisoning points. We show that this can provide a misleading analysis of the robustness of the algorithms, as the hyperparameters' values can change depending on the type and strength of the attack. We propose a novel and more general poisoning attack formulation to test ML algorithms that use hyperparameters in worst-case scenarios. In such cases, the attacker is aware of the dataset used, aims to maximize the overall error and the attack can be modelled as a \emph{bilevel optimization} problem where the outer problem is a \emph{minimax} that includes both the learning of the poisoning points and the hyperparameters and the inner problem involves learning the model's parameters. We have applied our formulation to test the robustness of Logistic Regression (LR) classifiers that use $L_2$ regularization. We have used \emph{hypergradient} (i.e. the gradient in the outer problem \citep{maclaurin2015gradient, franceschi2017forward, franceschi2018bilevel, grazzi2020iteration}) descent/ascent to solve the minimax bilevel optimization problem. Our experiments show that the results reported in \citep{xiao2015feature} provide an overly pessimistic  view of the effect of regularization. When used with the right regularization hyperparameter, $L_2$ regularization helps partially mitigate the effect of poisoning attacks. We show that the value of the regularization hyperparameter (if optimized) can increase with the attack strength, i.e., the algorithm tries to compensate the negative effect of the attack by increasing the strength of the regularization term.  
We start by setting the problem, study the effect of regularization in a small illustrative example and then present our findings.

\section{General Optimal Poisoning Attacks} \label{sec:generalAttacks}
In a classification task, given the input space $\mathcal{X}\subseteq \mathbb{R}^m$ and the label space $\mathcal{Y}$, the learner aims to estimate the mapping $f: \mathcal{X} \rightarrow \mathcal{Y}$. Given a training set $\mathcal{D}_\text{tr}= \{(\textbf{x}_{\text{tr}_i} , y_{\text{tr}_i})\}^{n_\text{tr}}_{i=1}$ drawn from the underlying
distribution $p(\mathcal{X}, \mathcal{Y})$, we can estimate $f$ with a model $\mathcal{M}$ trained by minimizing a loss function $\mathcal{L}(\mathcal{D}_\text{tr}, \boldsymbol{\Lambda}, {\textbf w})$ w.r.t. its parameters, $\textbf{w}\in \mathbb{R}^d$, given the hyperparameters $\boldsymbol{\Lambda}\in \mathbb{R}^c$. Similarly to previous work \citep{foo2008efficient, xiao2015feature}, we assume the learner has access to a small validation dataset $\mathcal{D}_\text{val}= \{(\textbf{x}_{\text{val}_j} , y_{\text{val}_j})\}^{n_\text{val}}_{j=1}$ with $n_\text{val}$ trusted points, representative of the underlying distribution. This validation set is held out for hyperparameter optimization, i.e. $\min_{\boldsymbol{\Lambda} \in \Phi(\boldsymbol{\Lambda})} \mathcal{L}(\mathcal{D}_\text{val}, \textbf{w}^\star)$, where $\textbf{w}^\star$ are the parameters learned and $\Phi(\cdot)$ is the feasible domain.

In data poisoning attacks the adversary injects a set of $n_\text{p}$ malicious data points, $\mathcal{D}_\text{p}= \{(\textbf{x}_{\text{p}_k} ,y_{\text{p}_k})\}^{n_\text{p}}_{k=1}$, in the training set to maximize  an objective function $\mathcal{A}$.  We assume the attacker can arbitrarily manipulate all the features and the label of the injected points, provided that they remain within the feasible domain of valid data points. We further assume the attacker knows everything about the training data, the feature representation, the loss function, and the ML model. This allows us to analyze worst-case scenario attacks of different strength. 

 In this work we focus on indiscriminate attacks, where the attacker aims to increase the overall classification error. Therefore, the attacker's objective function $\mathcal{A}$ coincides with the defender's loss function $\mathcal{L}$ evaluated on $\mathcal{D}_\text{val}$ at the end of training, i.e. $\mathcal{L}(\mathcal{D}_\text{val}, {\bf w}^{\star})$.  Hence, the attacker aims to maximize the loss evaluated on the defender's validation set, which can be formulated as a bilevel optimization problem where the outer objective is the attacker's objective, which depends on the optimal parameters obtained from the inner (training) problem. Previous studies have neglected the effect of the regularization hyperparameter in the attacker's problem \citep{biggio2012poisoning, xiao2015feature, mei2015using, munoz2017towards, koh2018stronger} and thus incompletely model the outer problem as a maximization w.r.t. the poisoning points. Taking hyperparameters into account, we propose to formulate the outer objective as a minimax problem: 
\begin{equation}
\begin{aligned}
\min_{\boldsymbol{\Lambda} \in \Phi(\boldsymbol{\Lambda})} &  \max_{\mathcal{D}_\text{p} \in \Phi(\mathcal{D}_\text{p})} \mathcal{L}(\mathcal{D}_\text{val}, \textbf{w}^\star) \\
\text{s.t.} &  \quad \textbf{w}^\star\in\argmin_{\textbf{w} \in \mathcal{W}}  \mathcal{L}\left(\mathcal{D}_\text{tr}', \boldsymbol{\Lambda}, \textbf{w}\right),\\
\end{aligned}
\label{eqAttacker2}
\end{equation} 
where  $\mathcal{D}_\text{tr}' = \mathcal{D}_\text{tr} \cup \mathcal{D}_\text{p}$ is the poisoned dataset. In the more general formulation we propose in (\ref{eqAttacker2}) it is clear that the poisoning points in $\mathcal{D}_\text{tr}'$ have an effect not only on the classifier's parameters, but also on its hyperparameters. This effect has previously been ignored.

Solving the bilevel problem in  (\ref{eqAttacker2}) is strongly NP-Hard \citep{bard2013practical} and the problem is, in general, non-convex. However, hypergradient-based approaches can be used to obtain (possibly) sub-optimal solutions, i.e. finding saddle points for the minimax problem in (\ref{eqAttacker2}). To compute the hypergradients for the outer objective, we assume that the loss function, $\mathcal{L}$, is convex and its first and second derivatives are Lipschitz-continuous. Similarly to \citep{biggio2012poisoning,mei2015using,xiao2015feature,munoz2017towards} we assume that the label of the poisoning points is set a priori, so the attacker just needs to learn their features ${\bf X}_\text{p}$. 

We use Reverse-Mode Differentiation (RMD) \citep{domke2012generic,maclaurin2015gradient,franceschi2017forward,munoz2017towards,franceschi2018bilevel, grazzi2020iteration} to estimate the hypergradients. RMD requires to first train for $T$ epochs. Then, the hypergradients estimate is computed by reversing the steps followed by the learning algorithm. Compared to grid search, this approach reduces the computational cost, as the algorithm does not need to be trained completely and evaluated for each combination of hyperparameters. After computing the hypergradients, we use projected hypergradient descent/ascent to update the poisoning points and the hyperparameters: 
\begin{equation}
  \begin{aligned}
{\bf X}_\text{p} & \leftarrow \Pi_{\Phi(\mathcal{D}_\text{p})} \left( {\bf X}_\text{p} + \alpha \ \nabla_{{\bf X}_\text{p}} \mathcal{A} \right), \hspace{1cm} {\boldsymbol \Lambda} & \leftarrow \Pi_{\Phi({\boldsymbol \Lambda})} \left( {\boldsymbol \Lambda} - \beta \ \nabla_{{\boldsymbol \Lambda}} \mathcal{A} \right),
  \end{aligned}
\label{eqUpdates}
\end{equation} where $\alpha$ and $\beta$ are respectively the learning rates and $\Pi_{\Phi(\mathcal{D}_\text{p})}$ and $\Pi_{\Phi({\boldsymbol \Lambda})}$ are the projection operators for the features of the poisoning points, ${\bf X}_\text{p}$, and the hyperparameters, ${\boldsymbol \Lambda}$.

\section{Illustrating the Mitigating Effect of \texorpdfstring{$L_2$}~~Regularization} \label{sec:L2}
$L_2$ (or Tikhonov) regularization is often used to increase the stability of ML algorithms \citep{xu2011sparse}. Using $L_2$ regularization we add a penalty term to the original loss function, which shrinks the norm of the model's parameters, so that $\mathcal{L}( \mathcal{D}_\text{tr}, \boldsymbol{\Lambda}, \textbf{w})=~\mathcal{L}(\mathcal{D}_\text{tr}, \textbf{w})+\frac{e^{\lambda}}{2}\left|\left|\textbf{w}\right|\right|_2^2,$
where $\lambda$ is the hyperparameter controlling the strength of the regularization term.

\begin{wrapfigure}{r}{.7\textwidth}
    \begin{minipage}{\linewidth}
    \centering
    \includegraphics[width=.32\textwidth]{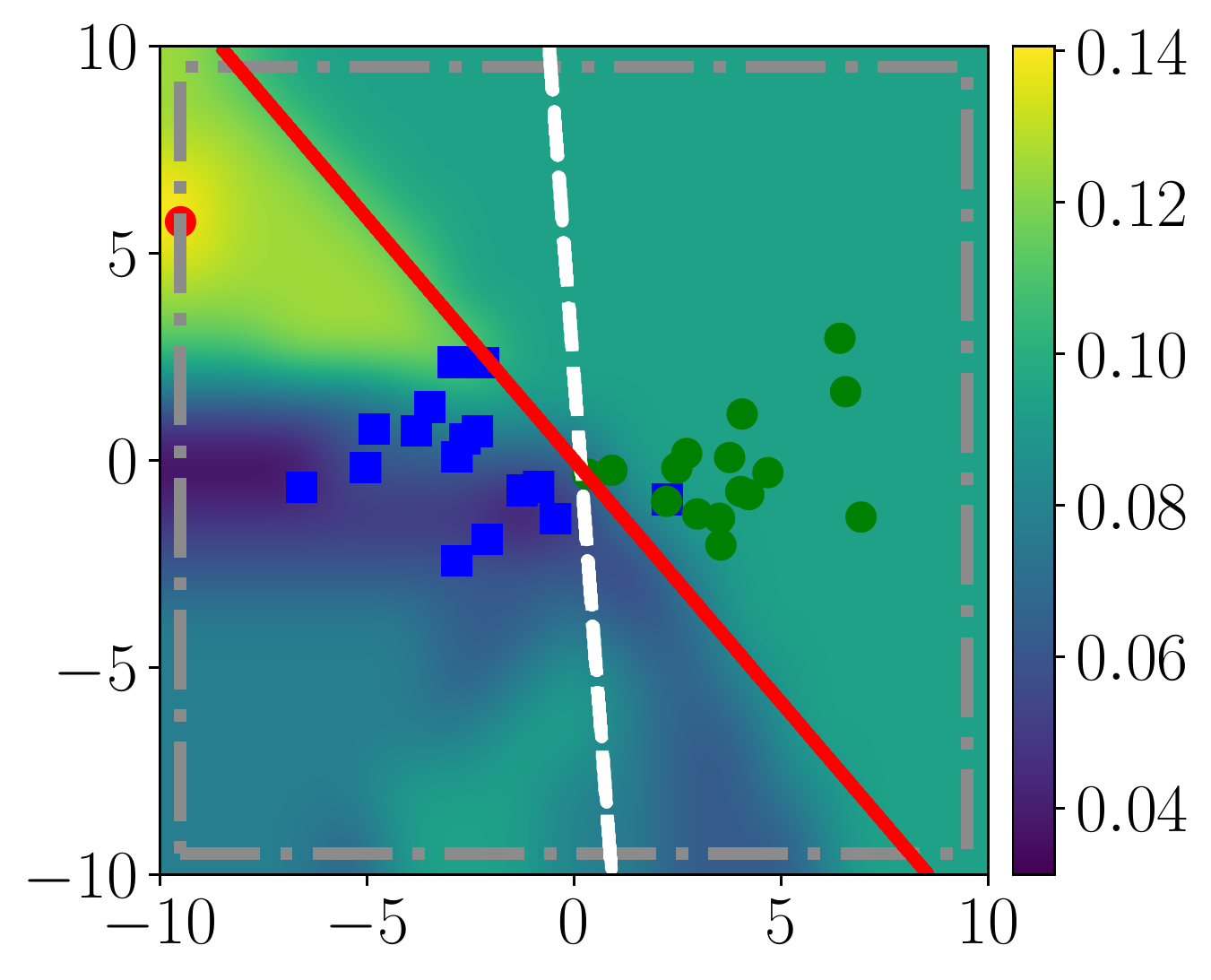}
    \includegraphics[width=.32\textwidth]{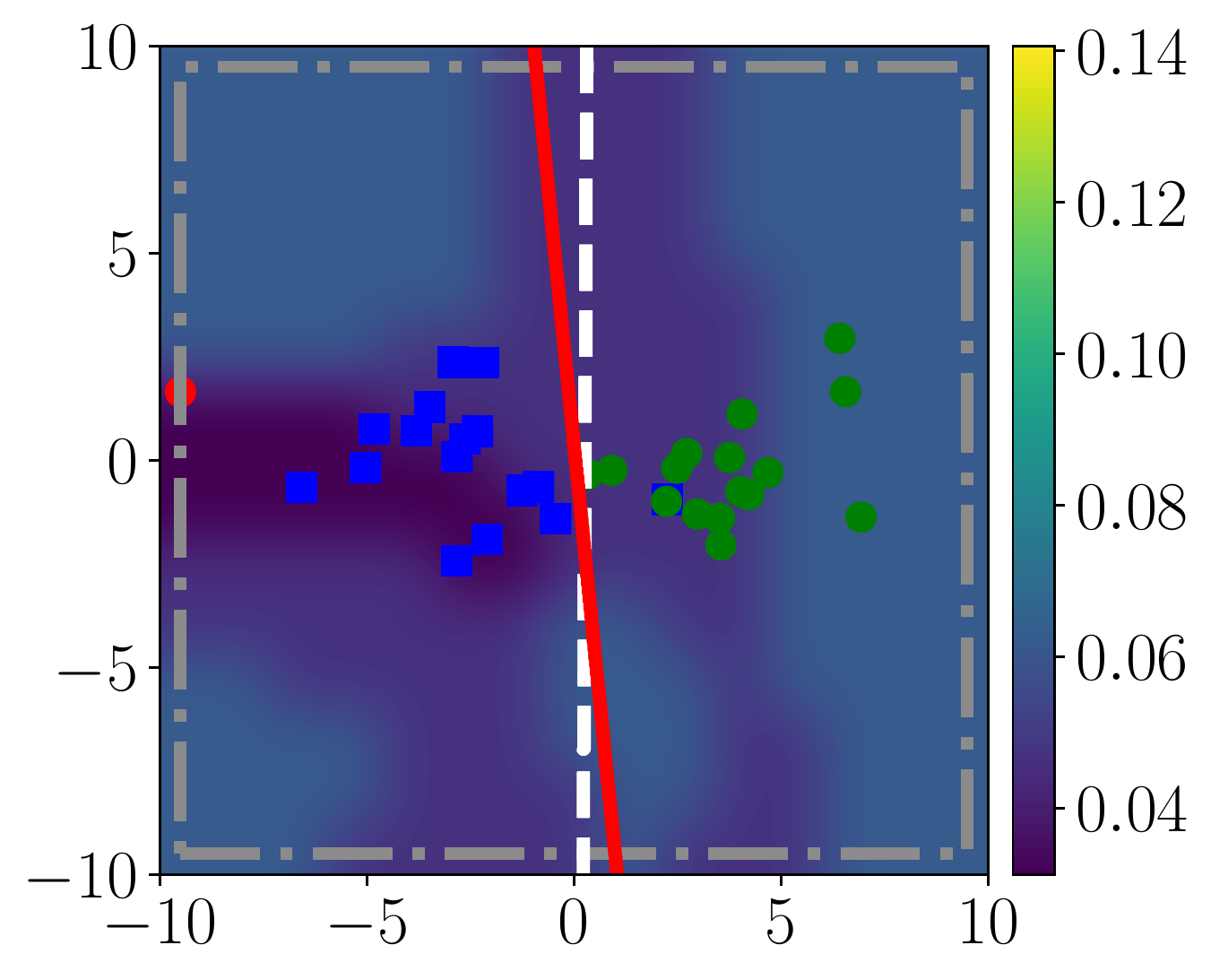}
    \includegraphics[width=.32\textwidth]{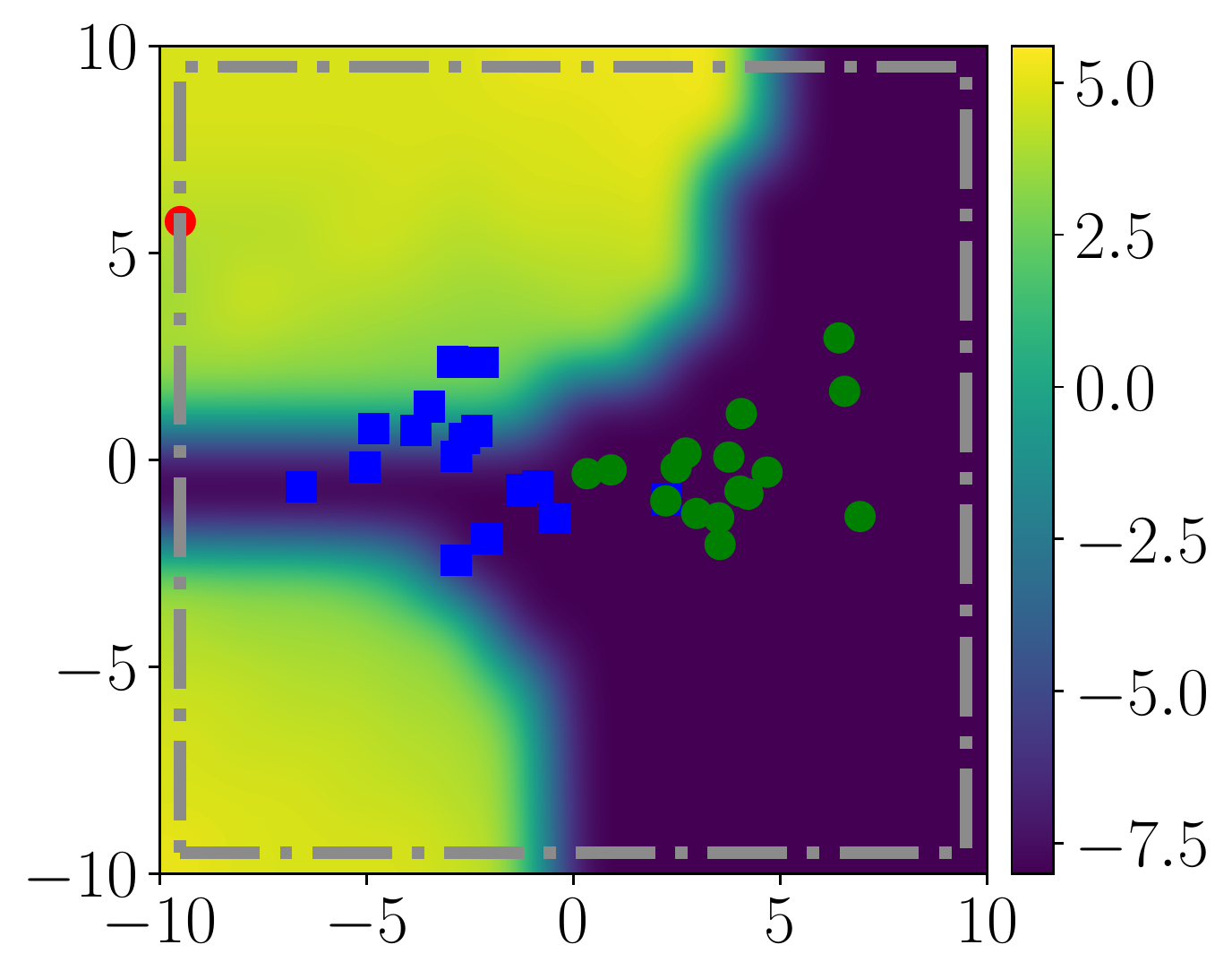}
\end{minipage}
\caption{Effect of regularization on a synthetic example with green and blue classes. The red point is the poisoning point (labelled as green). The dashed-dotted grey box represent the attacker's constraints. (Left) Standard LR with no regularization. (Center) LR with $L_2$ regularization ($\lambda\approx3$).  (Right) Value of $\lambda$ learned that minimizes the validation error as a function of the injected poisoning point.\vspace{-.27cm}}
	\label{fig:synthetic}
\end{wrapfigure}

In Fig.~\ref{fig:synthetic} we illustrate the effect of regularization and  the limitations of the approach in \citep{xiao2015feature} using a synthetic example with an LR classifier. Fig.~\ref{fig:synthetic}(left) shows the effect of injecting a single poisoning point that maximizes the error (measured on a separate validation set) against a non-regularized LR classifier. 
The dashed-white line represents the decision boundary learned when training on the clean dataset, whereas the red line depicts the decision boundary when training on the poisoned dataset. We can observe that a single poisoning point can significantly alter the decision boundary. Fig.~\ref{fig:synthetic}(center), shows a similar scenario, but training an LR classifier with $L_2$ regularization, setting a large value for $\lambda$ ($\lambda\approx3$). Here, we can observe that the effect of the poisoning is significantly reduced and the decision boundary shifts only slightly. In the background of Fig.~\ref{fig:synthetic}(left) and Fig.~\ref{fig:synthetic}(center) we represent the validation error of the LR trained on a poisoned dataset as a function of the location of the poisoning point. We can observe that, when there is no regularization (left) the error can significantly increase when we inject the poisoning point in certain regions. On the contrary, when regularization is applied (center), the colormap is more uniform, i.e. the algorithm is quite stable regardless of the position of the poisoning point.  

Fig.~\ref{fig:synthetic}(right) shows how the value of the regularization hyperparameter, $\lambda$, changes as a function of the poisoning point. The colormap in the background represents the value of $\lambda$ that minimizes the error on the validation set. We can observe that $\lambda$ changes significantly depending on the position of the poisoning point. Thus, $\lambda$ is much bigger for the regions where the poisoning point can influence the classifier more  (Fig.~\ref{fig:synthetic}(left)). Thus, when the poisoning attack can have a very negative impact on the classifier's performance, the importance of the regularization term, controlled by $\lambda$, increases. It is thus clear that selecting the value of $\lambda$ appropriately can have a significant impact on the classifier's robustness. Furthermore, when testing the robustness of $L_2$-regularized classifiers it is necessary to consider the interplay between the attack strength and the value of $\lambda$. 

\section{Experiments}
\label{sec:experiment}
\begin{figure*}[!t]
	\begin{centering}
		\begin{subfigure}[b]{0.27\textwidth}
			\includegraphics[width=\textwidth\vspace{-.12cm}]{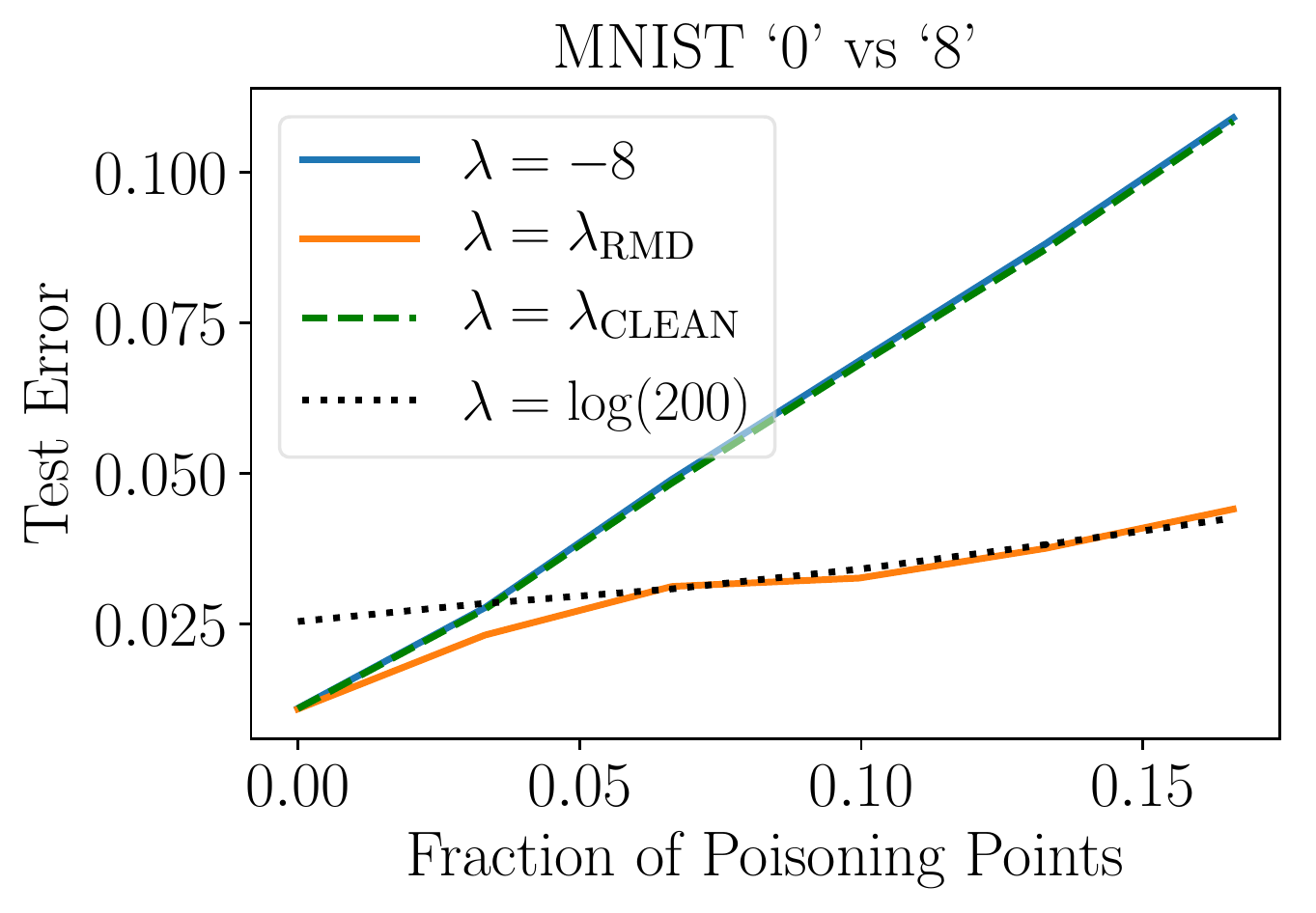}
			\caption{}
		\end{subfigure}
					\hspace{-.2cm}
		\enskip 
		\begin{subfigure}[b]{0.26\textwidth}
			\includegraphics[width=\textwidth\vspace{-.12cm}]{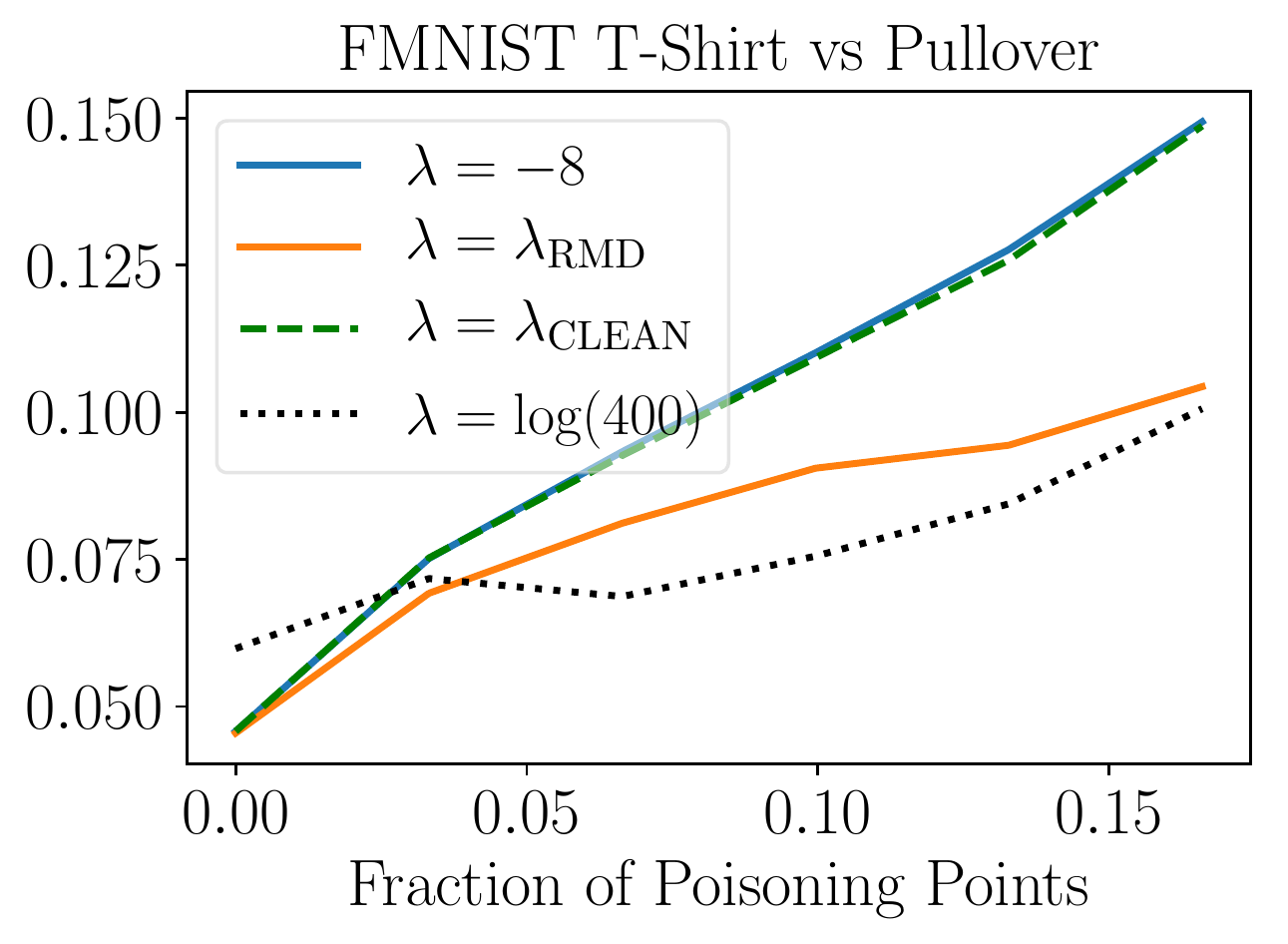}
			\caption{}
		\end{subfigure}
		\hspace{-.2cm}
		\enskip 
		\begin{subfigure}[b]{0.279\textwidth}
			\includegraphics[width=\textwidth\vspace{-.12cm}]{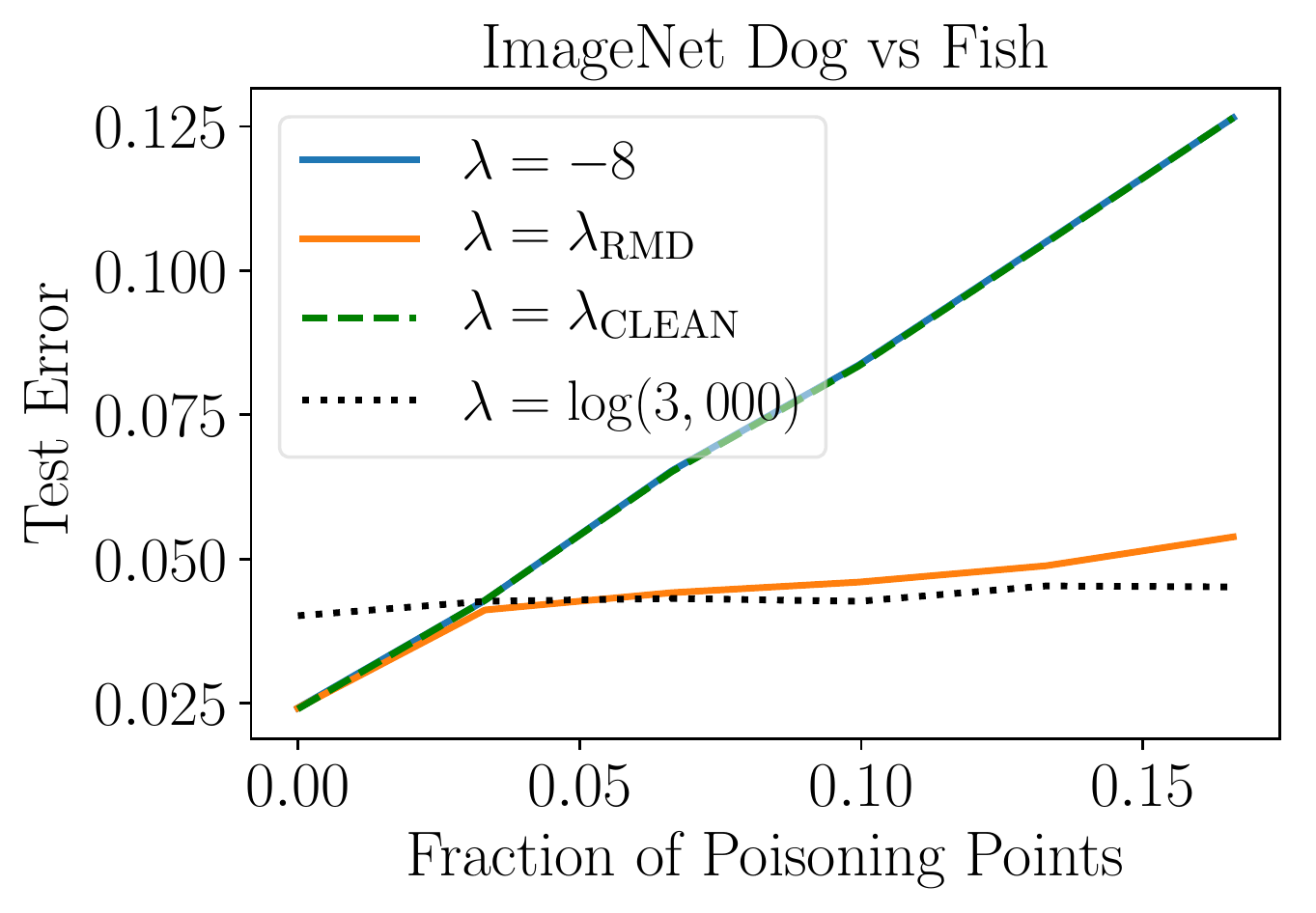}
			\caption{}
		\end{subfigure}
		\enskip 
		\begin{subfigure}[b]{0.28\textwidth}
			\includegraphics[width=\textwidth\vspace{-.12cm}]{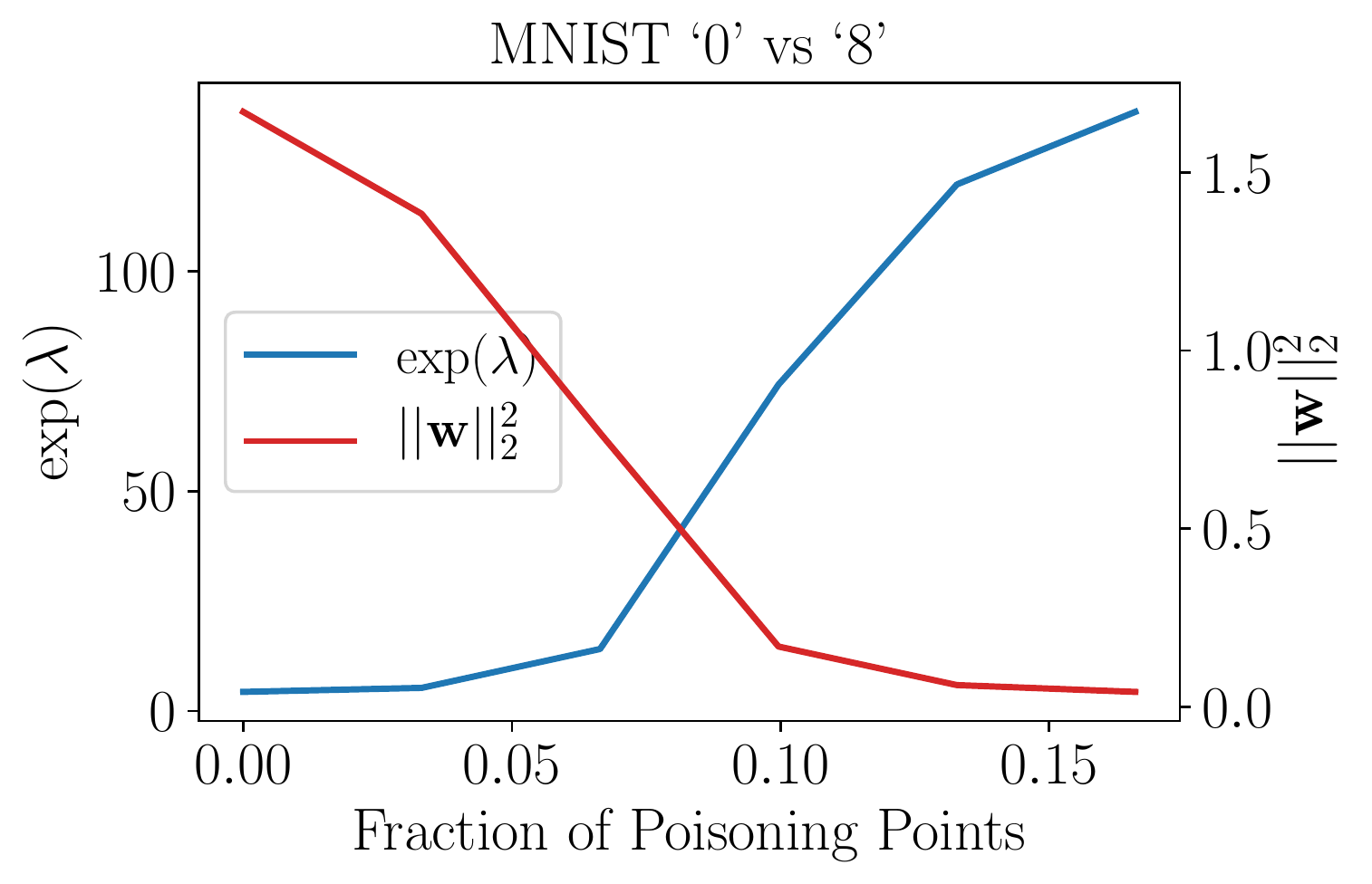}
			\caption{}
		\end{subfigure}
		\hspace{-.2cm}
		\enskip 
		\begin{subfigure}[b]{0.283\textwidth}
			\includegraphics[width=\textwidth\vspace{-.12cm}]{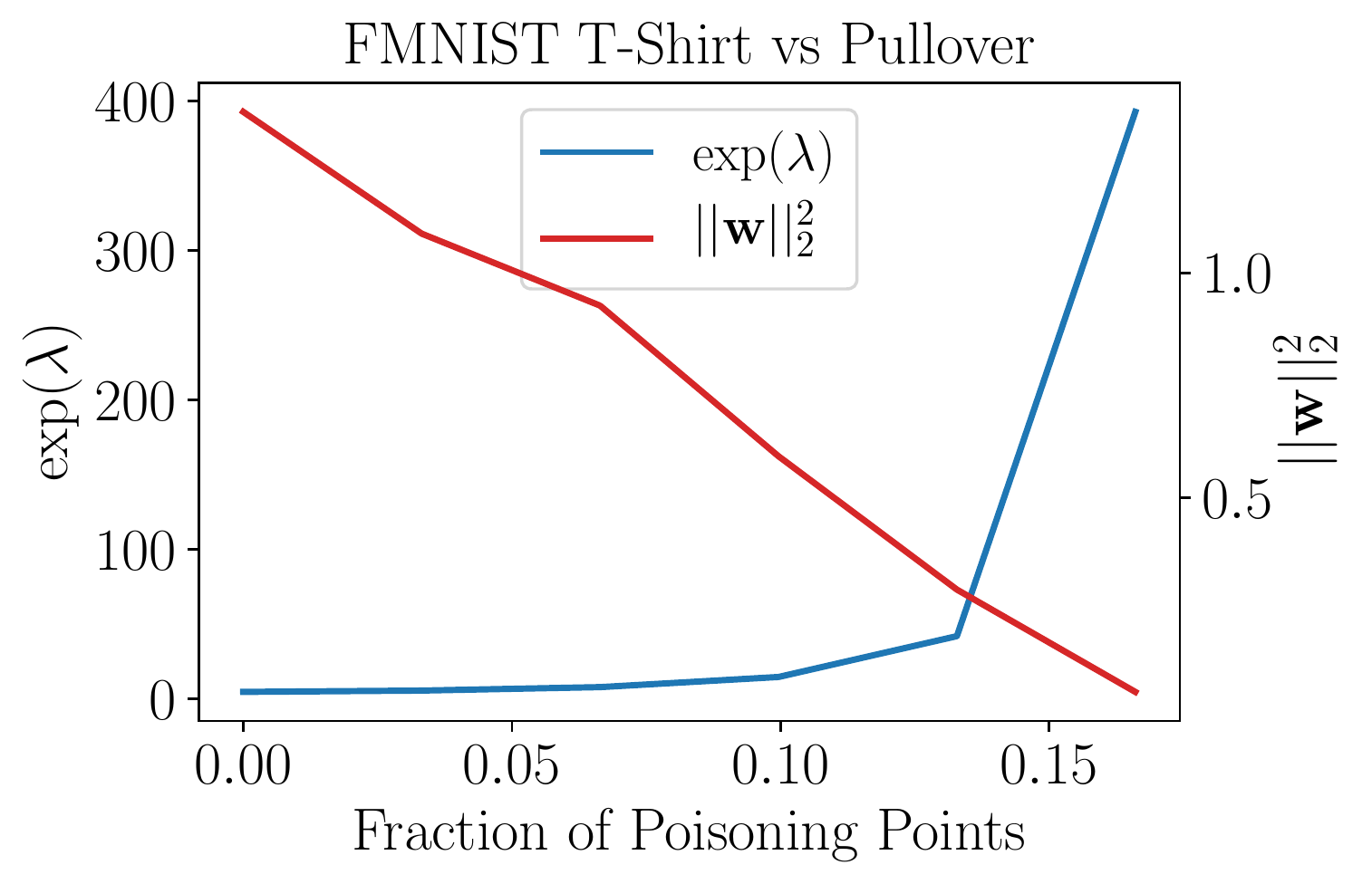}
			\caption{}
		\end{subfigure}
		\hspace{-.2cm}
		\enskip 
		\begin{subfigure}[b]{0.288\textwidth}
			\includegraphics[width=\textwidth\vspace{-.1cm}]{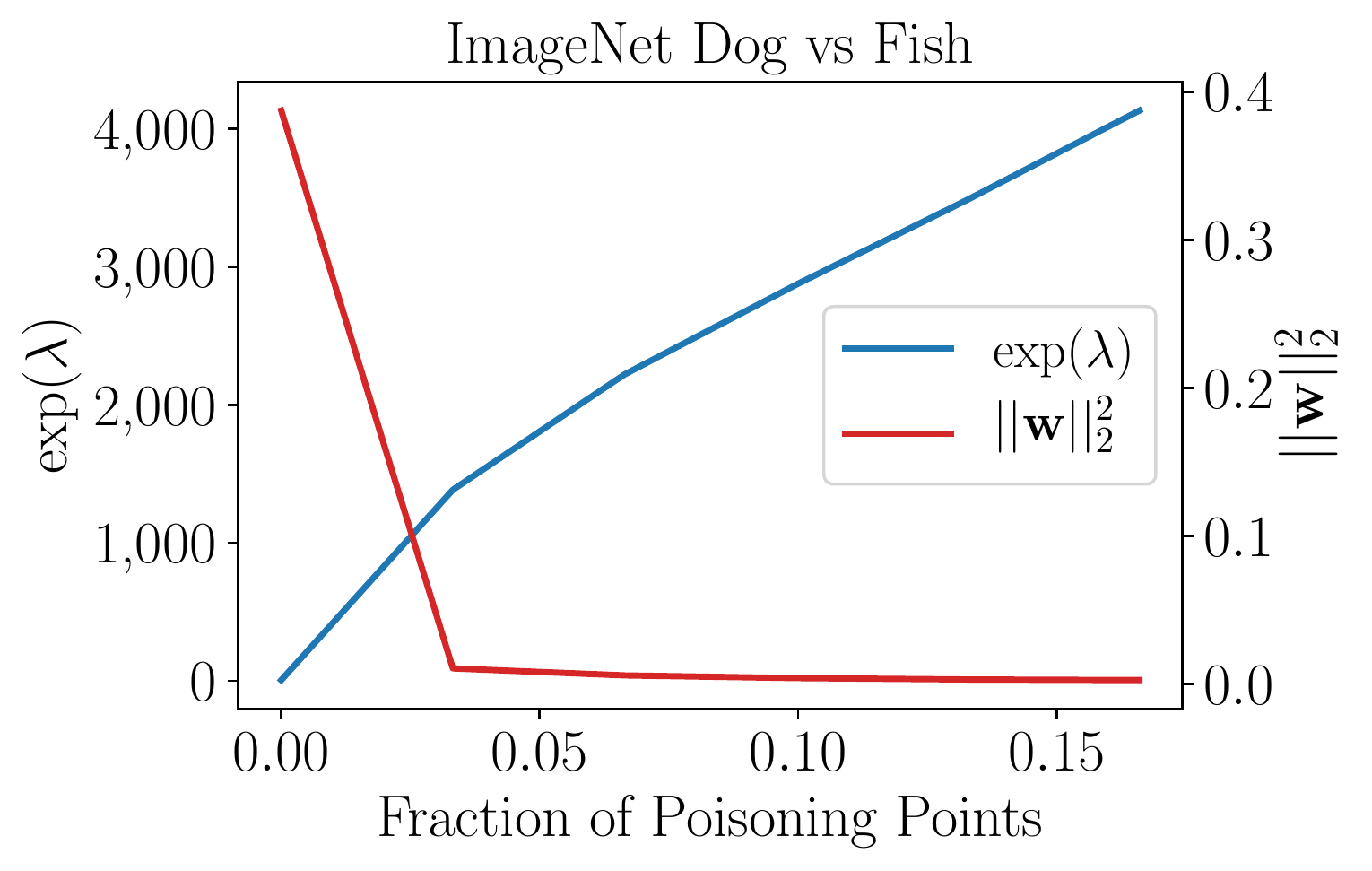}
			\caption{}
		\end{subfigure}
	\end{centering}
	\vspace{-.24cm}
	\caption{Results for the optimal attack against LR: The first row represents the average test error on (a) MNIST, (b) FMNIST, and (c) ImageNet. The second row contains the plots for the average $\lambda$ and $||\textbf{w}||_2^2$ for (d) MNIST, (e) FMNIST, and (f) ImageNet.\vspace{-.46cm}}
	\label{fig:lropt}
\end{figure*}
In this section, we evaluate the effectiveness of the attack strategy in (\ref{eqAttacker2}) against LR.\footnote{The PyTorch implementation of the algorithms used for the experiments is available at {\color{blue} \href{https://github.com/javiccano/regularization-and-poisoning---secml}{https://github.com/javiccano/regularization-and-poisoning---secml}}.} We use three different binary classification problems: MNIST (`0' vs `8') \citep{lecun1998gradient}, Fashion-MNIST (FMNIST) (\emph{T-shirt} vs \emph{pullover}) \citep{xiao2017fashion}, and ImageNet (\emph{dog} vs \emph{fish}) \citep{russakovsky2015imagenet}  preprocessed as in \citep{koh2017understanding} (we use the same Inception-v3 \citep{szegedy2016rethinking} features).\footnote{The details of the experimental setup  and hardware used for the experiments can be found in the Appendix.} We assume the attacker knows the value of $\lambda$ for the defender. To evaluate the effect of the regularization hyperparameter, we craft optimal poisoning attacks, setting the value of $\lambda$ to different constant values: a very small regularization term, a very large one, and the value of $\lambda$ optimized with $5$-fold cross-validation, training the model on the clean dataset ($\lambda_\text{CLEAN}$). The latter is similar to the settings used in \citep{xiao2015feature}.  
We compare these values with the value of $\lambda_\text{RMD}$ calculated according to Eq. (\ref{eqAttacker2}).

In Fig.~\ref{fig:lropt}(a)-(c) we can observe that for the small $\lambda$ and $\lambda_\text{CLEAN}$ the attacks are very effective and the test error increases significantly when compared to the algorithm's performance on the clean dataset. In contrast, for the largest $\lambda$ the test error increases only slightly with the increasing fraction of poisoning points, showing a stable performance regardless of the attack strength. However, in the absence of an attack, the algorithm clearly \emph{underfits} and the error is significantly higher compared to the other models. When the value of $\lambda$ is learned ($\lambda_\text{RMD}$) by solving the problem in (\ref{eqAttacker2}), the increase in the test error remains significantly smaller and the algorithm has a similar performance to the case where the value of $\lambda$ is large as the fraction of poisoning points increases. In this sense, the error does not increase further by injecting more poisoning points. In the absence of an attack, the performance for $\lambda_\text{RMD}$ is as good as in the case of $\lambda_\text{CLEAN}$. These results show that the attack and the methodology presented in \citep{xiao2015feature} provide an overly pessimistic view on the robustness of $L_2$ regularization to poisoning attacks and that, by appropriately selecting the value of $\lambda$, we can effectively reduce the impact of poisoning attacks. We can also observe that there is a trade-off between accuracy and robustness: setting a very large value for $\lambda$ makes the algorithm more robust, but the performance on clean data is degraded. Our formulation allows to learn the value of $\lambda$ most appropriate for the training set. 

In Fig.~\ref{fig:lropt}(d)-(f) we show the value of $\lambda$ learned and the norm of the model's parameters, $||{\bf w}||^2_2$, as a function of the fraction of poisoning points injected for the solution of problem (\ref{eqAttacker2}) with RMD. We can observe that in all cases, the regularization hyperparameter increases as we increase the fraction of poisoning points. This confirms that the regularization term tries to compensate the effect of the malicious samples on the model's parameters. Thus, as expected, $L_2$ provides a natural mechanism to stabilize the model in the presence of attacks. For ImageNet, the order of magnitude of $\lambda$ is higher, as there are more features in the model. In Fig.~\ref{fig:lropt}(d)-(f) we can also observe that, as expected from the properties of $L_2$ regularization, when $\lambda$ increases, the norm of the parameters decreases.

\section{Conclusions}
\label{sec:conclusion}
We have shown that previous approaches to optimal poisoning attacks, where the model's hyperparameters are considered constant, provide a misleading view of the algorithms' robustness. We have evidenced that poisoning attacks can have a strong influence on the hyperparameters learned, and thus, their influence must be considered when assessing algorithm robustness. To achieve this, we have introduced a novel worst-case poisoning attack strategy to evaluate the robustness of ML classifiers that contain hyperparameters. The attack is formulated as a minimax bilevel optimization problem that can be solved with hypergradient-based techniques. We have shown that choosing a fixed a priori value of $\lambda$ can be detrimental: if the value is too high it damages accuracy, if the value is too low it damages robustness. In this sense, in contrast to the results reported in \citep{xiao2015feature} (which uses a fixed regularization hyperparameter), when facing stronger attacks i.e., more poisoning points, the regularization hyperparameter can also increase to compensate the instability that the attacker tries to produce in the algorithm. It is therefore important for organizations training models from data that may be untrusted, to regularly adjust hyperparameters to adapt the performance and robustness of the algorithm to current threats.

\subsubsection*{Acknowledgments}
We gratefully acknowledge financial support for this work from the UK Defence Science and Technology Laboratory (Dstl); contract no: DSTLX-1000120987.

\bibliography{iclr2021_conference}
\bibliographystyle{iclr2021_conference}

\appendix
\section{Hessian-Vector Products}
\label{sec:hvp}
Let $\textbf{x}\in \mathcal{X} \subseteq \mathbb{R}^m$, $\textbf{y} \in \mathcal{Y} \subseteq \mathbb{R}^n$, $\textbf{v}\in\mathbb{R}^m$, and $f(\textbf{x},\textbf{y})\in\mathbb{R}$. If the second partial derivatives of $f(\textbf{x},\textbf{y})$ are continuous \emph{almost everywhere}\footnote{A property
that holds almost everywhere holds throughout all space except on a set of \emph{measure zero}. Intuitively, a set of measure zero occupies a negligible volume in the measured space \citep{goodfellow2016deep}.} in $\mathcal{X}$ and $\mathcal{Y}$ (mild condition), the Hessian-vector products $\left(\nabla_\textbf{x}^2f(\textbf{x},\textbf{y})\right)\textbf{v}$ and $\left(\nabla_\textbf{y}\nabla_\textbf{x}f(\textbf{x},\textbf{y})\right)\tran\textbf{v}$ can be computed exactly and efficiently by using the following identities  \citep{pearlmutter1994fast}:
\begin{gather}
  \begin{aligned}
\begin{split}
\left(\nabla_\textbf{x}^2f(\textbf{x},\textbf{y})\right)\textbf{v} &= \nabla_\textbf{x}\left(\textbf{v}\tran\nabla_\textbf{x}f(\textbf{x},\textbf{y})\right),\\
\left(\nabla_\textbf{y}\nabla_\textbf{x}f(\textbf{x},\textbf{y})\right)\tran\textbf{v} &= \nabla_\textbf{y}\left(\textbf{v}\tran\nabla_\textbf{x}f(\textbf{x},\textbf{y})\right).
\end{split}
  \end{aligned}
\end{gather}
The upper and lower expressions scale as $\mathcal{O}(m)$ and $\mathcal{O}(\max(m, n))$, respectively, both in time and space. Analogous expressions can be obtained when $\textbf{v}\in\mathbb{R}^n$. An elegant aspect of this technique is that, for machine learning models optimized with gradient-based methods, the equations for exactly evaluating
the Hessian-vector products emulate closely those for standard forward and backward propagation. Hence, the application of existing automatic differentiation frameworks to compute this product is typically straightforward \citep{pearlmutter1994fast, bishop2006pattern}. 

\section{Reverse-Mode Differentiation Algorithm}
As described in \citep{franceschi2017forward}, we can think of the training algorithm as a discrete-time dynamical system, described by a set of states ${\bf s}^{(t)} \in\mathbb{R}^{d_\text{s}}$, with $t = 0, \ldots, T$, where each state depends on model's parameters, the accumulated gradients and/or the velocities. In this paper, we focus on stochastic gradient descent, so that ${\bf s}^{(t)} = {\bf w}^{(t)}$. Then, from a reduced number of training iterations, $T$, we can estimate the hypergradients from the values of the parameters collected in the set of states. Depending on the order to compute the operations we can differentiate two approaches to estimate the hypergradients: Reverse-Mode (RMD) and Forward-Mode Differentiation (FMD) \citep{griewank2008evaluating,franceschi2017forward}. In the first case, RMD requires first to train the learning algorithm for $T$ epochs, and then, to compute ${\bf w}^{(0)}$ to ${\bf w}^{(T)}$. Then, the hypergradients estimate is computed by reversing the steps followed by the learning algorithm from ${\bf w}^{(T)}$ down to ${\bf w}^{(0)}$. On the other hand, FMD computes the estimate of the hypergradients as the algorithm is trained, i.e. from ${\bf w}^{(0)}$ to ${\bf w}^{(T)}$ (i.e. the estimates can be computed in parallel with the training procedure).

To estimate the hypergradients, RMD requires to compute a forward and a backward pass through the set of states. In contrast, FMD just needs to do the forward computation. However, the scalability of FMD depends heavily on the number of hyperparameters compared to RMD. Then, for problems where the number of hyperparameters is large, as is the case for the poisoning attacks we introduced in the paper, RMD is computationally more efficient to estimate the hypergradients. For this reason, we used RMD in our experiments.

Here we include the Reverse-Mode Differentiation (RMD) algorithm (Alg. \ref{alg:bg}) \citep{domke2012generic, maclaurin2015gradient, franceschi2017forward, munoz2017towards, franceschi2018bilevel, grazzi2020iteration}, which we use to compute the hypergradient estimate at the outer level problem (both for the features of the poisoning points, $\textbf{X}_\text{p}$, and the hyperparameters, $\boldsymbol{\Lambda}$).  We make use of a similar notation to the algorithms presented in \citep{maclaurin2015gradient, munoz2017towards}. The resulting algorithm for the case of $\boldsymbol{\Lambda}$ is analogous. 

On the other hand, the choice of the number of epochs for the inner problem, $T$, depends on the model and the training dataset. Low values of $T$ could lead to low-quality approximations for the hypergradient. As $T$ increases, the solution of RMD approaches the exact (true) hypergradient, but at the risk of overfitting the outer objective in the bilevel optimization problem \citep{franceschi2018bilevel}.

\begin{algorithm}[!t]
	\caption{Reverse-Mode Differentiation}
	\label{alg:bg}
	
	\begin{flushleft}
	
    {\bfseries Input:}  $\mathcal{M}$, $\mathcal{A}$, $\mathcal{L}$,  $\mathcal{D}_\text{val}$, $\mathcal{D}_\text{tr}'^{(\tau)}$, $\boldsymbol{\Lambda}^{(\tau)}$, $\textbf{w}^{(0)}$,  $T$, $\eta$ \\
		{\bfseries Output:} $\nabla_{\textbf{X}_\text{p}}\mathcal{A}\left(\textbf{w}^{(T)}\right)$
		
	\end{flushleft}
	
	\begin{algorithmic}[1]
		
		\FOR{$t=0$ \textbf{to} $T-1$}
		
		\STATE $\textbf{g} \leftarrow  \nabla_{\textbf{w}_{}}\mathcal{L}\left(\textbf{w}^{(t)}\right)$ 
		
		\STATE $\textbf{w}^{(t+1)}\leftarrow \textbf{w}^{(t)} - \eta \textbf{g}$ \COMMENT{stochastic gradient descent}
		\ENDFOR
		
		\STATE $d\textbf{X}_\text{p}^{(T)} \leftarrow \textbf{0}$
		
		\STATE $d\textbf{w}^{(T)} \leftarrow \nabla_{\textbf{w}_{}}\mathcal{A}\left(\textbf{w}^{(T)}\right)$
		\FOR{$t=T-1$ \textbf{down to} $0$}
		
		\STATE $d\textbf{g}_\text{w} \gets \left(\nabla^2_{\textbf{w}_{}}\mathcal{L}\left(\textbf{w}^{(t)}\right)\right)d\textbf{w}^{(t+1)} $ \COMMENT{Hessian-vector product (Sect. \ref{sec:hvp})}
		\STATE $d\textbf{g}_{\text{X}_\text{p}} \gets \left(\nabla_{\textbf{X}_\text{p}}\nabla_{\textbf{w}_{}}\mathcal{L}\left(\textbf{w}^{(t)}\right)\right)\tran d\textbf{w}^{(t+1)} $ \COMMENT{Hessian-vector product (Sect. \ref{sec:hvp})}
		
				\STATE	$
		d\textbf{w}^{(t)}  \leftarrow d\textbf{w}^{(t+1)}  - \eta d\textbf{g}_\text{w}
		$
		
		\STATE	$
		d\textbf{X}_\text{p}^{(t)}  \leftarrow d\textbf{X}_\text{p}^{(t+1)}  - \eta d\textbf{g}_{\text{X}_\text{p}}
		$

		\ENDFOR
		\STATE $ \nabla_{\textbf{X}_\text{p}}\mathcal{A}\left(\textbf{w}^{(T)}\right)
		\leftarrow d\textbf{X}_\text{p}^{(0)}$

	\end{algorithmic}
\end{algorithm}

\section{Experimental Settings}
In the following specifications, $T_{\mathcal{D}_\text{p}}$ denotes the number of hyperiterations at the outer problem when the poisoning points are learned and $\lambda$ is fixed, $T_\lambda$ when the training set is clean and $\lambda$ is learned, and $T_\text{mul}$ when $\textbf{X}_\text{p}$ and $\lambda$ are learned in a coordinate manner. Finally, $\eta_\text{tr}$ is the learning rate when testing the attack.
\label{sec:expset}
\subsection{Synthetic Example}
For the synthetic experiment shown in Sect.~4 of the paper, we sample the attacker's data from two bivariate Gaussian distributions, $\mathcal{N}(\boldsymbol{\mu}_0, \boldsymbol{\Sigma}_0)$ and $\mathcal{N}(\boldsymbol{\mu}_1, \boldsymbol{\Sigma}_1)$, with mean vectors $\boldsymbol{\mu}_i$ and covariance matrices $\boldsymbol{\Sigma}_i$ for each class, $i=0,1$:
\begin{equation*}
  \begin{aligned}
    \boldsymbol{\mu}_0 & = \begin{bmatrix} -3.0 \\ \hspace{\minuslength}0.0 \end{bmatrix}, & \qquad \boldsymbol{\Sigma}_0 & = \begin{bmatrix} 2.5 & 0.0 \\ 0.0 & 1.5 \end{bmatrix}, \\
    \boldsymbol{\mu}_1  & = \begin{bmatrix} 3.0 \\ 0.0 \end{bmatrix}, &         \boldsymbol{\Sigma}_1 & = \begin{bmatrix} 2.5 & 0.0 \\ 0.0 & 1.5 \end{bmatrix}. \\
  \end{aligned}
\end{equation*}
The attacker uses $32$ points ($16$ per class) for training and $64$ ($32$ per class) for validation, and one poisoning point cloned from the validation set (in the example of the paper, cloned from the set labelled as blue), whose label is flipped. This poisoning point is concatenated into the training set and the features of this point are optimized with RMD. In order to poison the LR classifier, we use a learning rate and a number of hyperiterations for the poisoning point $\alpha=0.4$, $T_{\mathcal{D}_\text{p}}=50$; the feasible domain $\Phi(\mathcal{D}_\text{p})\in[-9.5,9.5]^2$; the learning rate and number of epochs for the inner problem $\eta=0.2$, $T=500$; and when evaluating the attack, $\eta_\text{tr}=0.2$, batch size $=32$ (full batch), and $\text{number of epochs} = 100$. When we apply regularization, we fix $\lambda=\log(20)\approx3$.

To plot the colormap that shows the value of the regularization hyperparameter learned (Fig. 1(right) of the paper) as a function of the poisoning point injected in the training set, the values of $\lambda$ explored for each possible poisoning point are in the range $[-8, 6]$. Then, the optimal value of $\lambda$ is chosen such that it minimizes the error of the model, trained on each combination of the poisoning point (concatenated into the training set) and $\lambda$ in the grid, and evaluated on the validation set.

\subsection{MNIST, FMNIST and ImageNet}
\begin{table*}[t]
	\centering
	\caption{Characteristics of the datasets used in the experiments.}{ 
		\begin{tabular}{|l|c|c|c|c|}
			\hline
			Name & \makecell{\# Training \\ Samples} & \makecell{\# Validation \\ Samples} & \makecell{\# Test \\ Samples} & \# Features \\
			\hline
			MNIST (`0' vs `8') & $512$ & $171$ & $1,954$ & $784$ \\
			FMNIST (\emph{T-shirt} vs \emph{pullover}) & $512$ & $171$ & $2,000$ & $784$ \\
			ImageNet (\emph{dog} vs \emph{fish}) & $512$ & $171$ & $600$ & $2,048$ \\
			\hline
	\end{tabular}}
	\label{tabDatasets}

\end{table*}

\begin{samepage}

\begin{table*}[!t]
	\centering
	\caption{Experimental settings for the poisoning attack.}{ 
		\begin{tabular}{|l|c|c|c|c|c|}
			\hline
			Name &  $T_{\mathcal{D}_\text{p}}$ &  $T_\lambda$ &  $T_\text{mul}$ &  $\alpha$ & $\beta$ \\
			\hline
			MNIST (`0' vs `8') (LR) & $100$ & $50$ & $100$ & $0.99$ & $0.80$ \\
			FMNIST (\emph{T-shirt} vs \emph{pullover}) (LR) & $100$ & $50$ & $100$ & $0.90$ & $0.30$ \\
			ImageNet (\emph{dog} vs \emph{fish}) (LR) & $100$ & $50$ & $200$ & $0.90$ & $0.40$ \\
			\hline
	\end{tabular}
	
	\vspace{0.2cm}
	
			\begin{tabular}{|l|c|c|c|c|}
			\hline
			Name & $\Phi(\mathcal{D}_\text{p})$ & $\Phi(\lambda)$ & $\eta$ & $T$  \\
			\hline
			MNIST (`0' vs `8') (LR) & $[0.0, 1.0]^{784}$ & $[-8,\log(200)]$ & $0.10$ & $150$ \\
			FMNIST (\emph{T-shirt} vs \emph{pullover}) (LR) & $[0.0, 1.0]^{784}$ & $[-8,\log(400)]$ & $0.08$ & $200$\\
			ImageNet (\emph{dog} vs \emph{fish}) (LR) & $[-0.5, 0.5]^{2,048}$ & $[-8,\log(20,000)]$ & $0.05$ & $200$ \\
			\hline
	\end{tabular}}
	\label{tabAttack}
\end{table*}

\begin{table*}[!t]
	\centering
	\caption{Experimental settings for training the models.}{ 
		\begin{tabular}{|l|c|c|c|}
			\hline
			Name &  $\eta_\text{tr}$ &  Batch Size &  Number of Epochs \\
			\hline
			MNIST (`0' vs `8') (LR) & $10^{-2}$ & $64$ & $200$ \\
			FMNIST (\emph{T-shirt} vs \emph{pullover}) (LR) & $10^{-2}$ & $64$ & $200$ \\
			ImageNet (\emph{dog} vs \emph{fish}) (LR) & $10^{-3}$ & $64$ & $300$ \\
			\hline
	\end{tabular}}
	\label{tabTrain}
\end{table*}
\end{samepage}
In our experiments, all the results for MNIST `0' vs `8' \citep{lecun1998gradient}, Fashion-MNIST (FMNIST) \emph{T-shirt} vs \emph{pullover} \citep{xiao2017fashion}, and ImageNet \emph{dog} vs \emph{fish} \citep{russakovsky2015imagenet} (preprocessed as in \citep{koh2017understanding}) are the average of $10$ repetitions with different random data splits for training and validation, whereas the test set is fixed. For all the attacks we measure the average test error for different attack strengths, where the number of poisoning points is in the range between $0$ and $85$.  To reduce the computational cost, the size of the batch of poisoning points that are simultaneously optimized
is $17$ for all the datasets. This way, we simulate six different ratios of poisoning ranging from $0\%$ to $16.6\%$.

The details of each dataset are included in Table \ref{tabDatasets}. All the datasets are balanced. Moreover, both MNIST and FMNIST sets are normalized to be in the range $[0, 1]^{784}$, whereas for the ImageNet sets we use the same Inception-v3 \citep{szegedy2016rethinking} features as in \citep{koh2017understanding}, and normalized them with respect to their training mean and standard deviation.

For all the experiments, we make use of stochastic gradient descent both to update the parameters in the forward pass of RMD\footnote{To refine the solutions obtained, when the poisoning points are learned and $\lambda$ is fixed, and when the training set is clean and $\lambda$ is learned, if the optimization algorithm gets stuck in a poor local optimum we restart the bilevel optimization procedure. In the case of the poisoning points, we reinitialize them with different values uniformly sampled without duplicates from the validation set. For the regularization hyperparameters, let $\lambda^{(\tau)}$ denote their latest value: We reinitialize them with values uniformly sampled from the range $\left[\lambda^{(\tau)}-0.5, \lambda^{(\tau)}+0.5\right]$. It is however noteworthy that these reinitializations were not crucial to obtain the results shown in the paper. The exploration of warm-restart techniques in the case of minimax bilevel problems is left for future work.} (full batch training), and to train the model when testing the attack (mini-batch training). The details of the attack settings are shown in Table \ref{tabAttack}, whereas the ones for training are in Table \ref{tabTrain}. Additionally, $\lambda_\text{CLEAN}$ is optimized with $5$-fold cross-validation---training the model on the clean dataset, as in \citep{xiao2015feature}. For all the datasets the ranges of values of $\lambda$ explored to compute $\lambda_\text{CLEAN}$ is $[-8, 1]$ (no better performance was observed for larger values of $\lambda$).  On the other hand, to accelerate the optimization of $\lambda_\text{RMD}$, when the training set is clean these hyperparameters are warm-started with a value $\lambda = \log(5)\approx1.61$.

\subsection{Details of the Hardware Used}
All the experiments are run on $2 \times 11$~GB NVIDIA GeForce\textregistered \hspace{0cm}  GTX 1080 Ti GPUs. The RAM memory is $64$~GB ($4\times16$~GB) Corsair VENGEANCE DDR4 $3000~\text{MHz}$. The processor (CPU) is Intel\textregistered \hspace{0cm} Core\texttrademark \hspace{0cm} i7 Quad Core Processor i7-7700k ($4.2$~GHz) $8$~MB Cache.

\section{Histograms of the Models' Parameters}

\begin{figure*}[!t]
	\begin{subfigure}[b]{0.316\textwidth}
		\includegraphics[width=\textwidth]{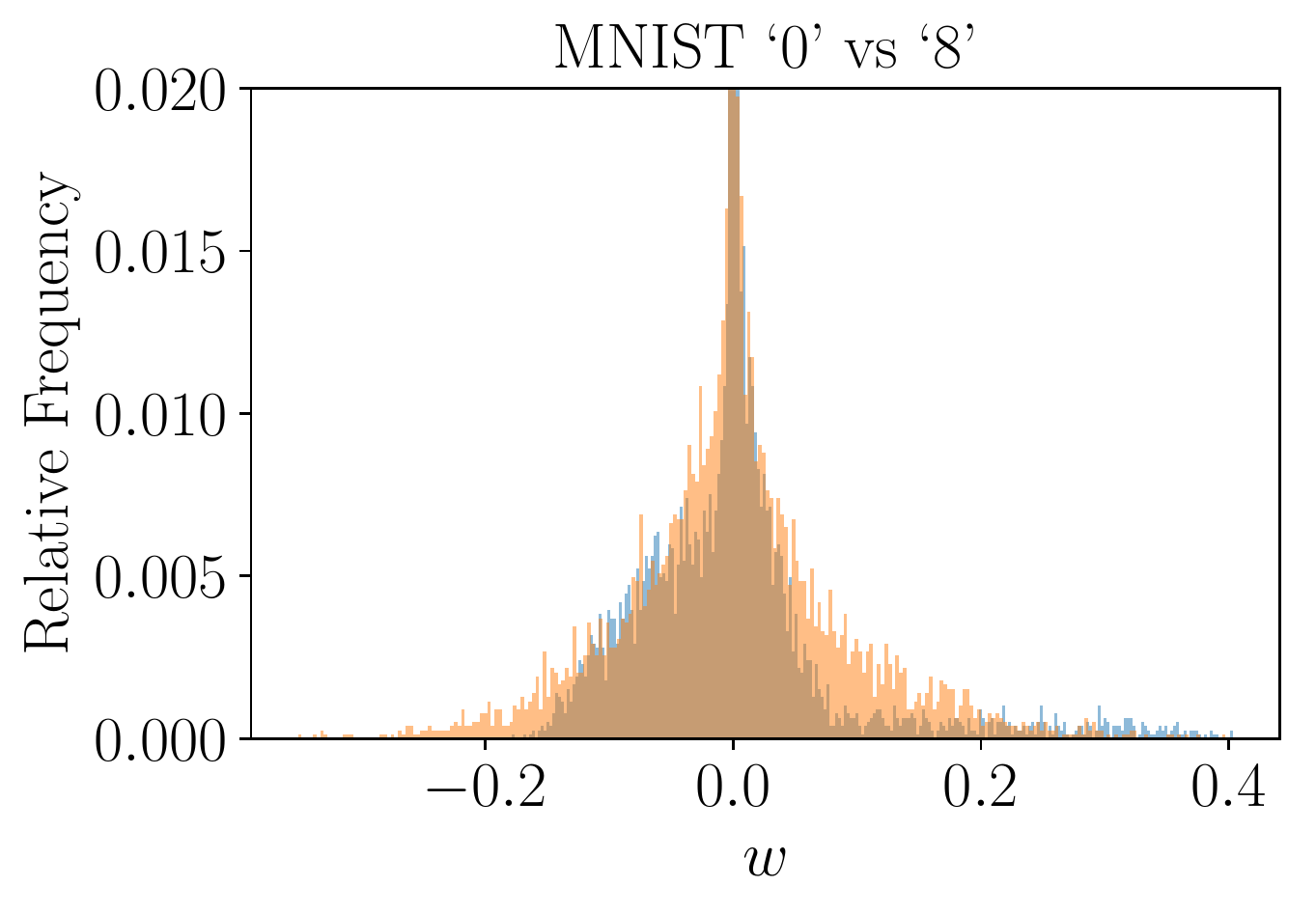}
		\caption{}
	\end{subfigure}
	\enskip 
	\begin{subfigure}[b]{0.295\textwidth}
		\includegraphics[width=\textwidth]{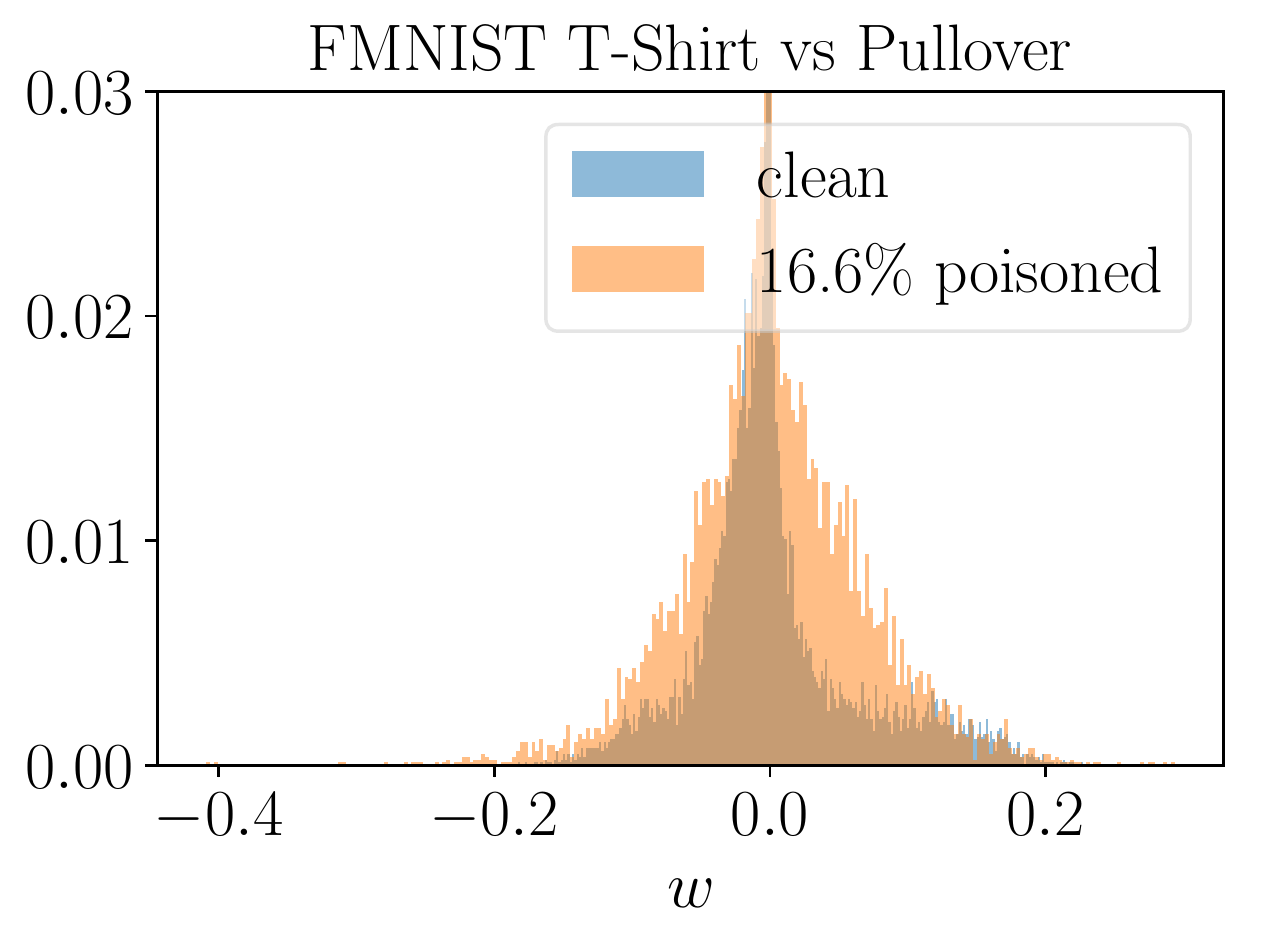}
		\caption{}
	\end{subfigure}
	\enskip 
	\begin{subfigure}[b]{0.319\textwidth}
		\includegraphics[width=\textwidth]{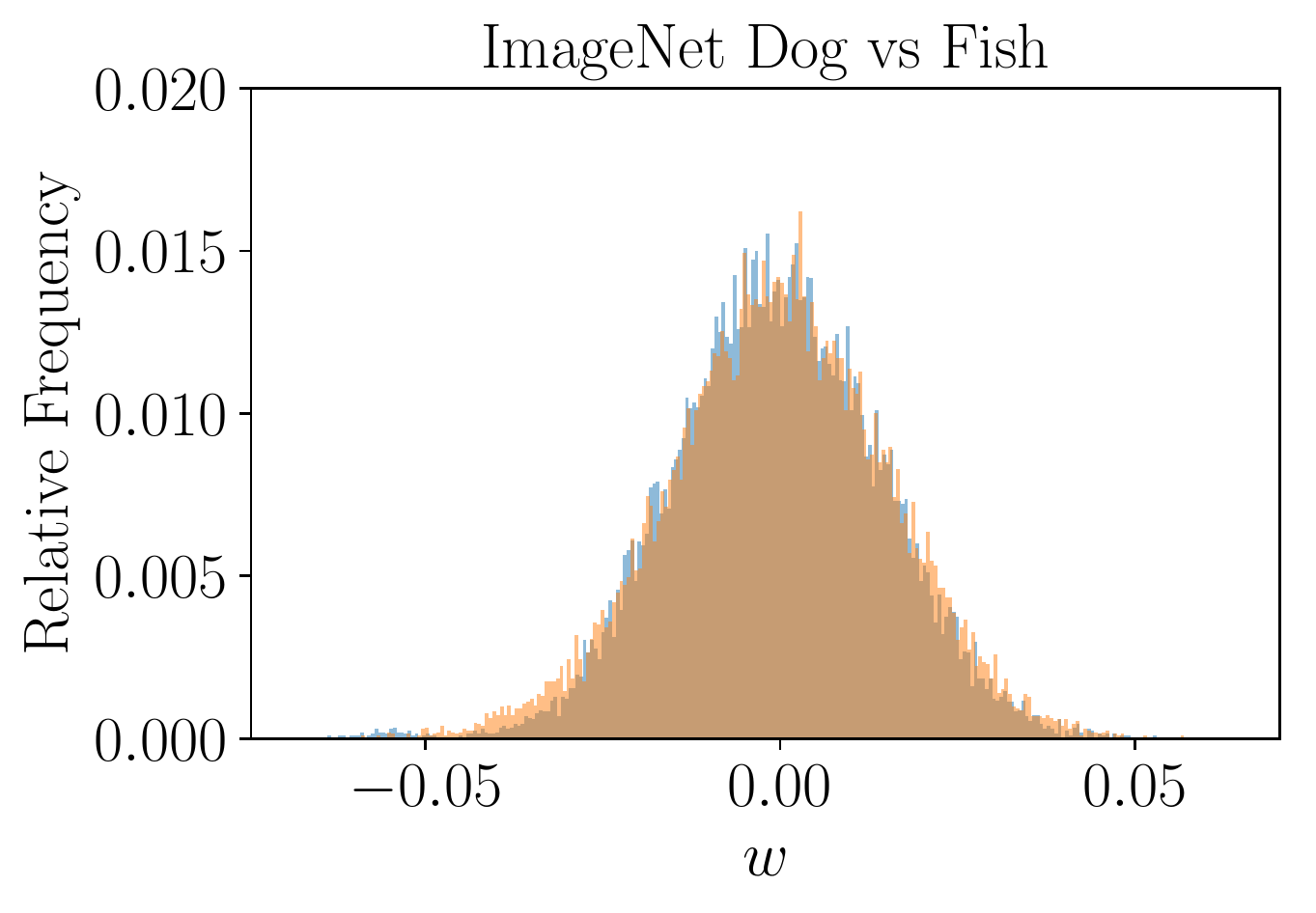}
		\caption{}
	\end{subfigure}
	\enskip 
	\begin{subfigure}[b]{0.323\textwidth}
		\includegraphics[width=\textwidth]{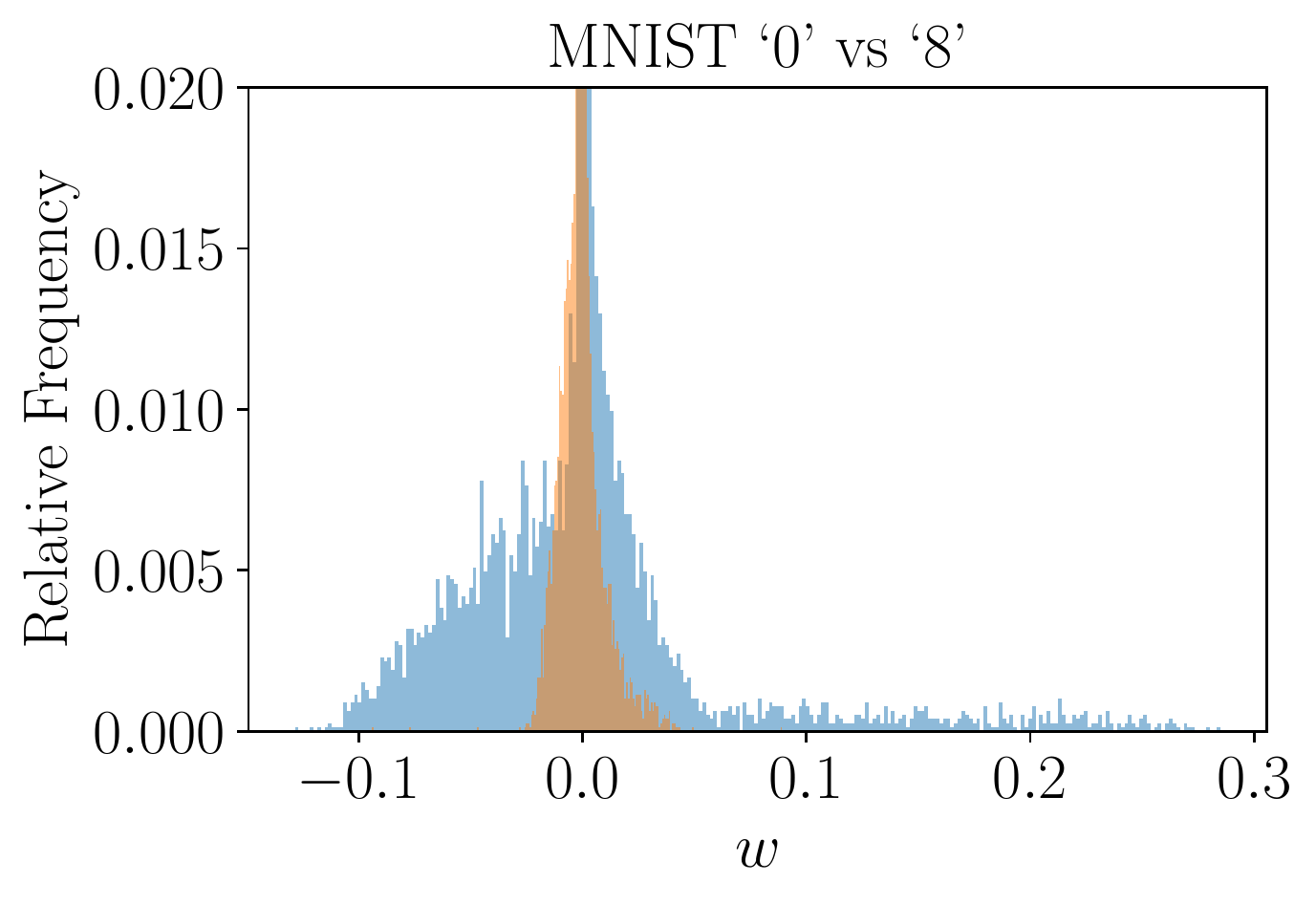}
		\caption{}
	\end{subfigure}
	\enskip 
	\begin{subfigure}[b]{0.3\textwidth}
		\includegraphics[width=\textwidth]{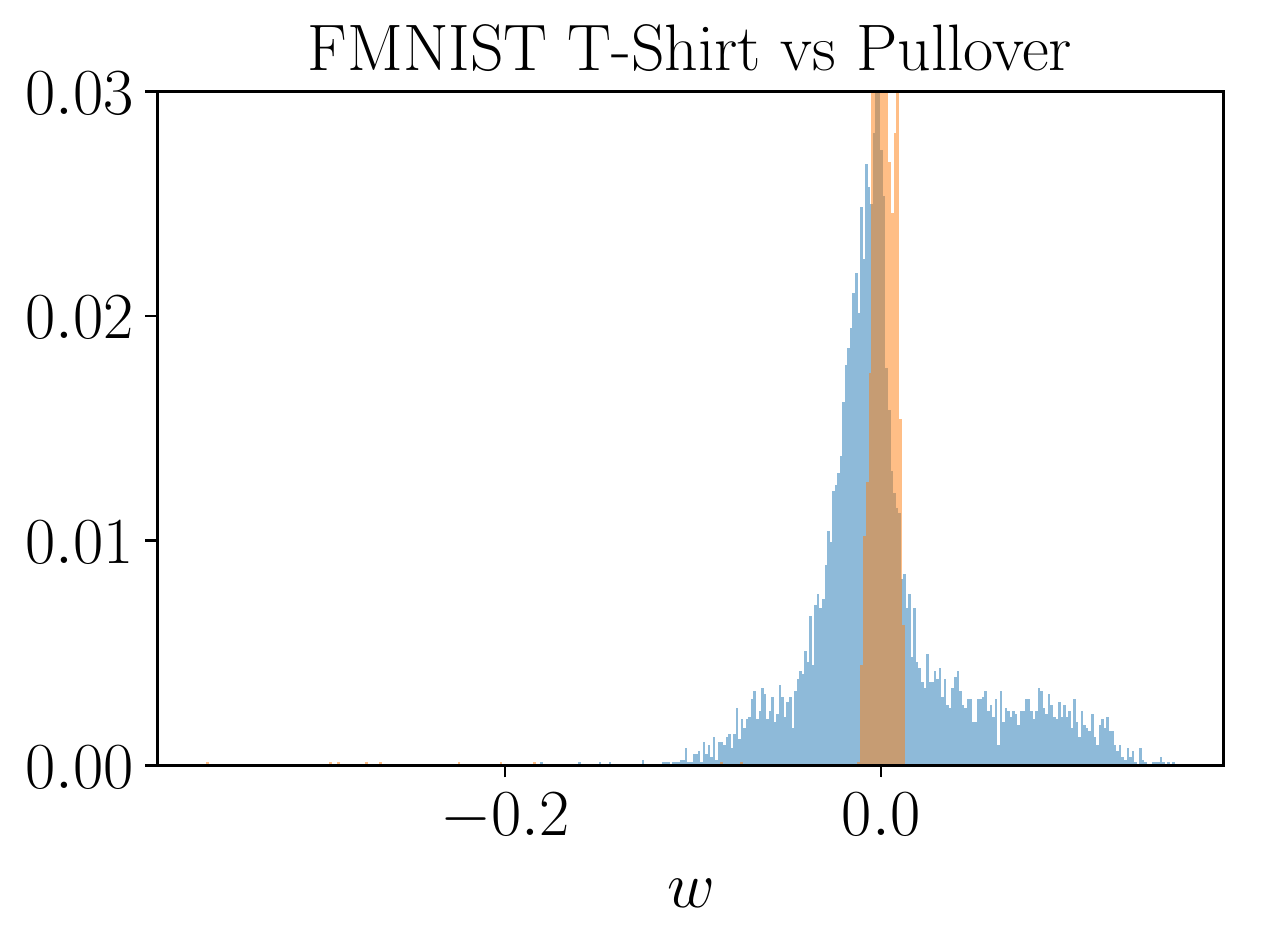}
		\caption{}
	\end{subfigure}
	\enskip 
	\begin{subfigure}[b]{0.323\textwidth}
		\includegraphics[width=\textwidth]{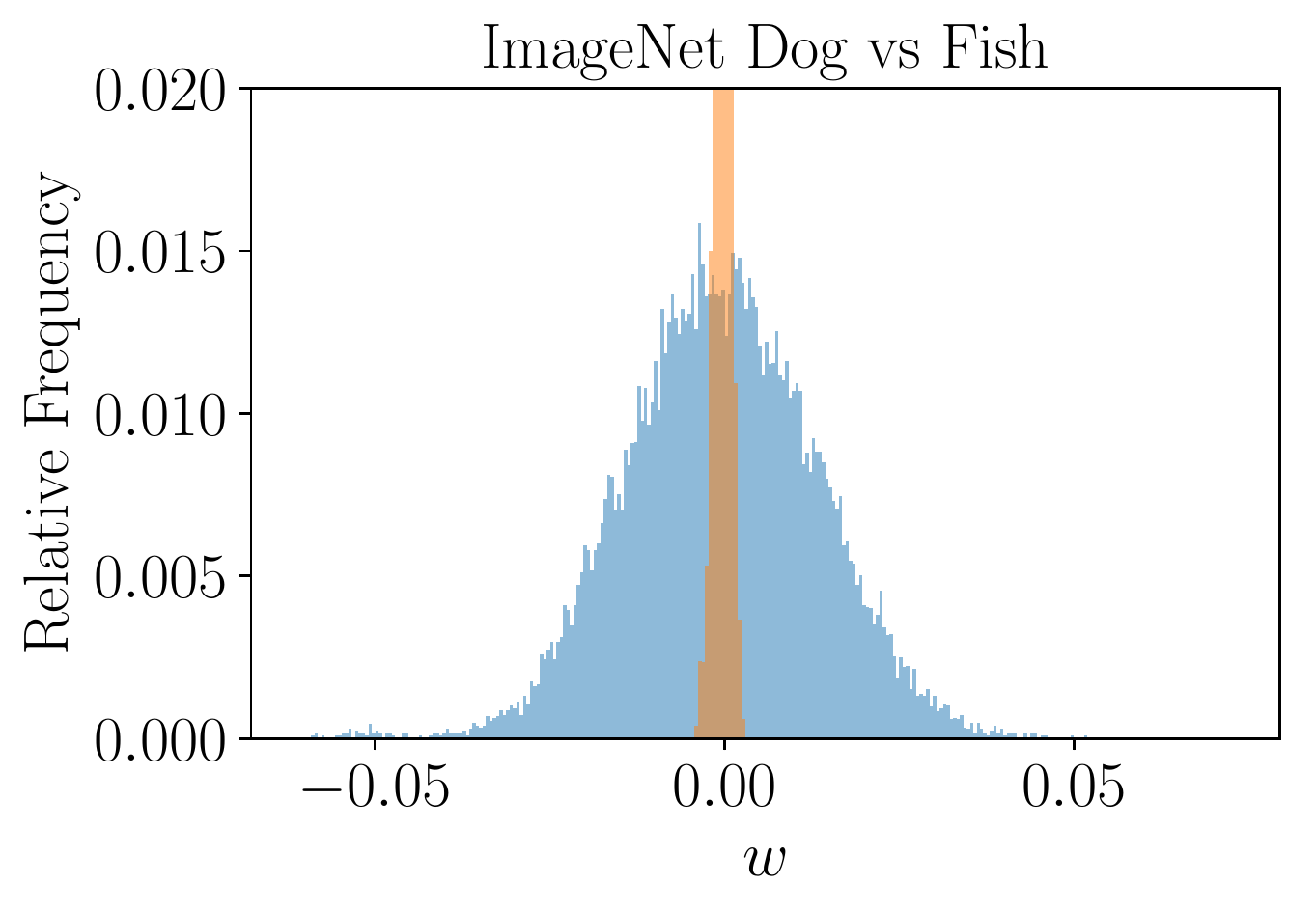}
		\caption{}
	\end{subfigure}
	\caption{Histograms (in terms of relative frequency) of the LR's parameters. For a better visualisation, the upper part of some subfigures has been omitted. The first row represents the case where no regularisation is applied: (a) MNIST, (b) FMNIST, and (c) ImageNet. The second row shows the case where $\lambda$ is learned with RMD: (d) MNIST, (e) FMNIST, and (f) ImageNet.}
	\label{fig:lrhist}
\end{figure*}

In Fig. \ref{fig:lrhist} we represent the histogram of the values of the parameters, $w$, of the LR classifier on MNIST, FMNIST and ImageNet, when the training sets are clean (blue bars) and when a $16.6\%$ of each training set is replaced by poisoning points  (orange bars). To appreciate better the distribution of the parameters, we omit the upper part of some plots. The first row depicts the cases when no regularisation is applied, and the second row shows the case where $\lambda$ is learned using RMD. We can clearly appreciate the effect of the regularisation:
For all the datasets, the range of values of the parameters (blue bars of the second row) is narrowed down, and when the attacker injects poisoning points, this forces the model to compress more these values (orange bars of the second row) close to $0$, as the value of $\lambda$ increases. This leads to a more stable model under possible malicious manipulations of the training data.

\end{document}